\documentclass[conference, 10pt]{IEEEtran}
\IEEEoverridecommandlockouts
\usepackage{cite}
\usepackage{amsmath,amssymb,amsfonts}
\usepackage{algorithm}
\usepackage{algorithmic}
\usepackage{graphicx}
\usepackage{textcomp}
\usepackage{xcolor}
\def\BibTeX{{\rm B\kern-.05em{\sc i\kern-.025em b}\kern-.08em
    T\kern-.1667em\lower.7ex\hbox{E}\kern-.125emX}}
\usepackage{soul}
\usepackage{enumitem}
\usepackage[super]{nth}
\usepackage{tikz}
\newcommand*\circled[1]{\tikz[baseline=(char.base)]{
		\node[shape=circle,draw,inner sep=0.2pt] (char) {#1};}}
\usepackage{fancyhdr}
\fancypagestyle{firstpage}{
  \fancyhf{}
  \chead{Accepted at the 62nd Design Automation Conference (DAC), June 2025, San Francisco, CA, USA.}
  \cfoot{\thepage}
}

\begin{document}

\title{Replay4NCL: An Efficient Memory Replay-based Methodology for Neuromorphic Continual Learning in Embedded AI Systems
\vspace{-0.7cm}
}

\author{
Mishal Fatima Minhas$^{1}$, Rachmad Vidya Wicaksana Putra$^{2}$, Falah Awwad$^{1}$, Osman Hasan$^{3}$, Muhammad Shafique$^{2}$ \\
$^{1}$\textit{Electrical and Communication Engineering Department,} \textit{United Arab Emirates University (UAEU), Al Ain, UAE}\\
$^{2}$\textit{eBrain Lab, New York University (NYU) Abu Dhabi, Abu Dhabi, UAE} \\
$^{3}$\textit{School of Electrical Engineering and Computer Science (SEECS), National University of Sciences and} \\ \textit{Technology (NUST), Islamabad, Pakistan} \\
\{mishal.fatima, f\_awwad\}@uaeu.ac.ae, \{rachmad.putra, muhammad.shafique\}@nyu.edu, \\ osman.hasan@seecs.edu.pk
\vspace{-0.5cm}}

\maketitle
\pagestyle{plain}
\thispagestyle{firstpage}

%%%%%%%%%%%%%%%%%%%%%%%%%%%%%%%%%%%%%%%%%%%%%%%%%%%%%%%%%%%%%%%%%%%%%%%%%%%%%%%%%%%%%%%%%
%%%%%%%%%%%%%%%%%%%%%%%%%%%%%%%%%%%%%%%%%%%%%%%%%%%%%%%%%%%%%%%%%%%%%%%%%%%%%%%%%%%%%%%%%

\begin{abstract} 
Neuromorphic Continual Learning (NCL) paradigm leverages Spiking Neural Networks (SNNs) to enable continual learning (CL) capabilities for AI systems to adapt to dynamically changing environments. 
Currently, the state-of-the-art employ a memory replay-based method to maintain the old knowledge. 
However, this technique relies on long timesteps and compression-decompression steps, thereby incurring significant latency and energy overheads, which are not suitable for tightly-constrained embedded AI systems (e.g., mobile agents/robotics). 
To address this, we propose \textit{Replay4NCL}, a novel efficient memory replay-based methodology for enabling NCL in embedded AI systems. 
Specifically, Replay4NCL compresses the latent data (old knowledge), then replays them during the NCL training phase with small timesteps, to minimize the processing latency and energy consumption. 
To compensate the information loss from reduced spikes, we adjust the neuron threshold potential and learning rate settings.
Experimental results on the class-incremental scenario with the Spiking Heidelberg Digits (SHD) dataset show that Replay4NCL can preserve old knowledge with Top-1 accuracy of 90.43\% compared to 86.22\% from the state-of-the-art, while effectively learning new tasks, achieving 4.88x latency speed-up, 20\% latent memory saving, and 36.43\% energy saving.
These results highlight the potential of our Replay4NCL methodology to further advances NCL capabilities for embedded AI systems. 
\end{abstract}


%%%%%%%%%%%%%%%%%%%%%%%%%%%%%%%%%%%%%%%%%%%%%%%%%%%%%%%%%%%%%%%%%%%%%%%%
\vspace{-0.2cm}
\section{Introduction}
\label{Sec_Intro}

\begin{figure}[t]
\centering
\includegraphics[width=\linewidth]{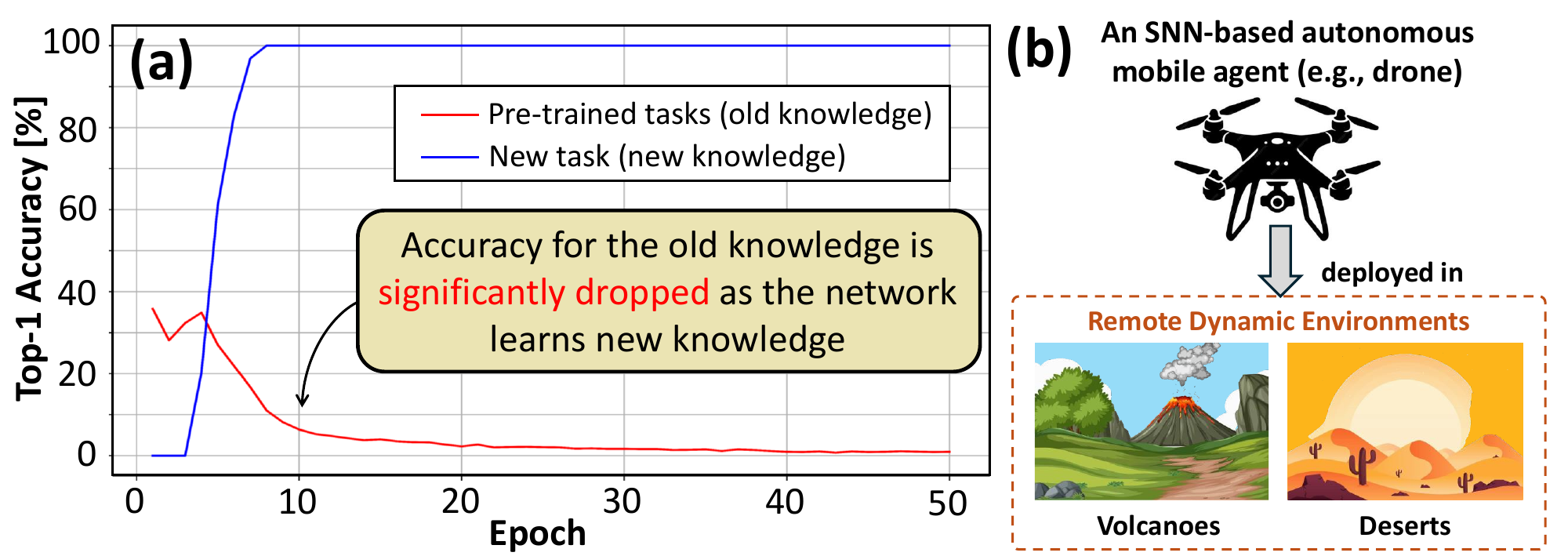}
\vspace{-0.7cm}
\caption{(a) Accuracy of the baseline network that faces CF issues in the class-incremental learning scenario. Baseline network refers to an SNN model without any NCL capabilities. (b) Illustration of the use-case of an embedded AI system that requires energy-efficient CL capabilities, e.g., a mobile agent.}
\label{Fig_CFissues}
\vspace{-0.6cm}
\end{figure}

The continual learning (CL) capability shown by human has inspired the developments of CL for artificial intelligence (AI) systems, e.g., through Deep Neural Networks (DNNs).
CL capability is highly desired by systems that are supposed to adapt to dynamically changing environments~\cite{bib122, bib19, bib42}, since the knowledge acquired during the off-line training (pre-training) phase may become outdated over time. 
This condition can result in reduced accuracy at run-time. 
To address this, the knowledge update is required to ensure correct functionality of the system.
However, simply performing the retraining technique from scratch is not practical, as it requires huge training time and energy consumption. 
Besides, it also requires a full dataset including data for the new task, which is not always available due to data restriction policy~\cite{Ref_Putra_RescueSNN_FNINS23}. 
Moreover, if the retraining is not performed carefully, it may also cause the \textit{catastrophic forgetting} (CF) issue, where accuracy for the previously trained tasks significantly decrease as the system learns a new task~\cite{bib15, bib16, bib61, bib44 , bib318, Ref_Lee_OvercomeCF_NIPS17}, as illustrated in Fig.~\ref{Fig_CFissues}(a).

Toward this, many researches have proposed different CL methods for DNNs~\cite{bib19}. 
However, the intensive DNN operations make it challenging to efficiently enable CL capabilities, especially in the application use-cases where embedded AI systems with limited battery capacity and resources are deployed in environments that are too far or not safe for humans (e.g., mobile agents/robots like UAVs and UGVs)~\cite{Ref_Putra_Mantis_ICARA23}, as shown in Fig.~\ref{Fig_CFissues}(b).

In recent years, \textit{neuromorphic continual learning} (NCL) has emerged as an attractive approach to enable efficient CL capabilities, i.e., by leveraging Spiking Neural Networks (SNNs) to perform sparse spike-based CL algorithms~\cite{Ref_Minhas_SurveyNCL_arXiv24, Ref_Panda_ASP_JETCAS18, Ref_Allred_CFN_FNINS20, Ref_Putra_SpikeDyn_DAC21, Ref_Putra_lpSpikeCon_IJCNN22, Ref_Davies_Loihi_MM18}. 
However, the NCL field is still very new, thereby practical solutions and methodologies for enabling efficient NCL capabilities in embedded AI systems is still not comprehensively explored.  
Therefore, \textbf{we mainly target the following research problem} in this paper, \textit{how can we enable efficient NCL capabilities for resource-constrained embedded AI systems?}
An efficient solution to this problem potentially enables the efficient NCL deployments for diverse applications use-cases.  

%%%%%%%%%%%%%%%%%%%%%%%%%
\subsection{State-of-the-Art and Their Limitations}

To address CF with NCL paradigm, previous works can be categorized into two different approaches based on the learning settings, i.e., \textit{unsupervised-} and \textit{supervised-based learning}. 

In the unsupervised learning settings, previous works have proposed different techniques, such as enhancements on bio-plausible Spike-Timing Dependent Plasticity (STDP) learning rules~\cite{Ref_Panda_ASP_JETCAS18, Ref_Allred_CFN_FNINS20, Ref_Putra_SpikeDyn_DAC21, Ref_Putra_lpSpikeCon_IJCNN22} and predictive coding~\cite{bib342}.
However, these studies still consider static and non-event-based datasets in their evaluations, e.g., MNIST~\cite{Ref_LeCun_MNIST_IEEE98} and Fashion MNIST~\cite{Ref_Xiao_FashionMNIST_arXiv17}.
Therefore, their applicability for event-based neuromorphic data (e.g., from dynamic vision sensors) is still not evaluated.

Meanwhile, in the supervised learning settings, recent works have also proposed different techniques, such as active dendrites~\cite{bib317}, architectural enhancements~\cite{bib304}~\cite{bib300}, memory replay~\cite{bib301}\cite{bib302}, regularization~\cite{bib339}, as well as Bayesian~\cite{bib299} and Hebbian~\cite{bib350} learning.
Currently, the state-of-the-art work employs the memory replay-based technique to preserve high accuracy on the old knowledge (i.e., \textit{latent knowledge}) while also learning new knowledge~\cite{bib302}; so-called the \textit{spiking latent replay (SpikingLR)} for brevity. 
However, this technique relies on long timesteps and compression-decompression steps to achieve high accuracy~\cite{bib302}, hence incurring significant latency and energy consumption overheads, which are not suitable for tightly-constrained embedded AI systems. 
These limitations indicate that \textit{an alternate efficient memory replay-based technique is highly needed to provide NCL capabilities with low memory, latency, and energy overheads}.

To illustrate the limitations of state-of-the-art and the optimization challenges, we perform an experimental case study, which is discussed further in Section~\ref{Sec_Intro_Challenges}. 

%%%%%%%%%%%%%%%%%%%%%%%%%
\subsection{Case Study and Associated Research Challenges}
\label{Sec_Intro_Challenges}

\begin{figure}[t]
\centering
\includegraphics[width=\linewidth]{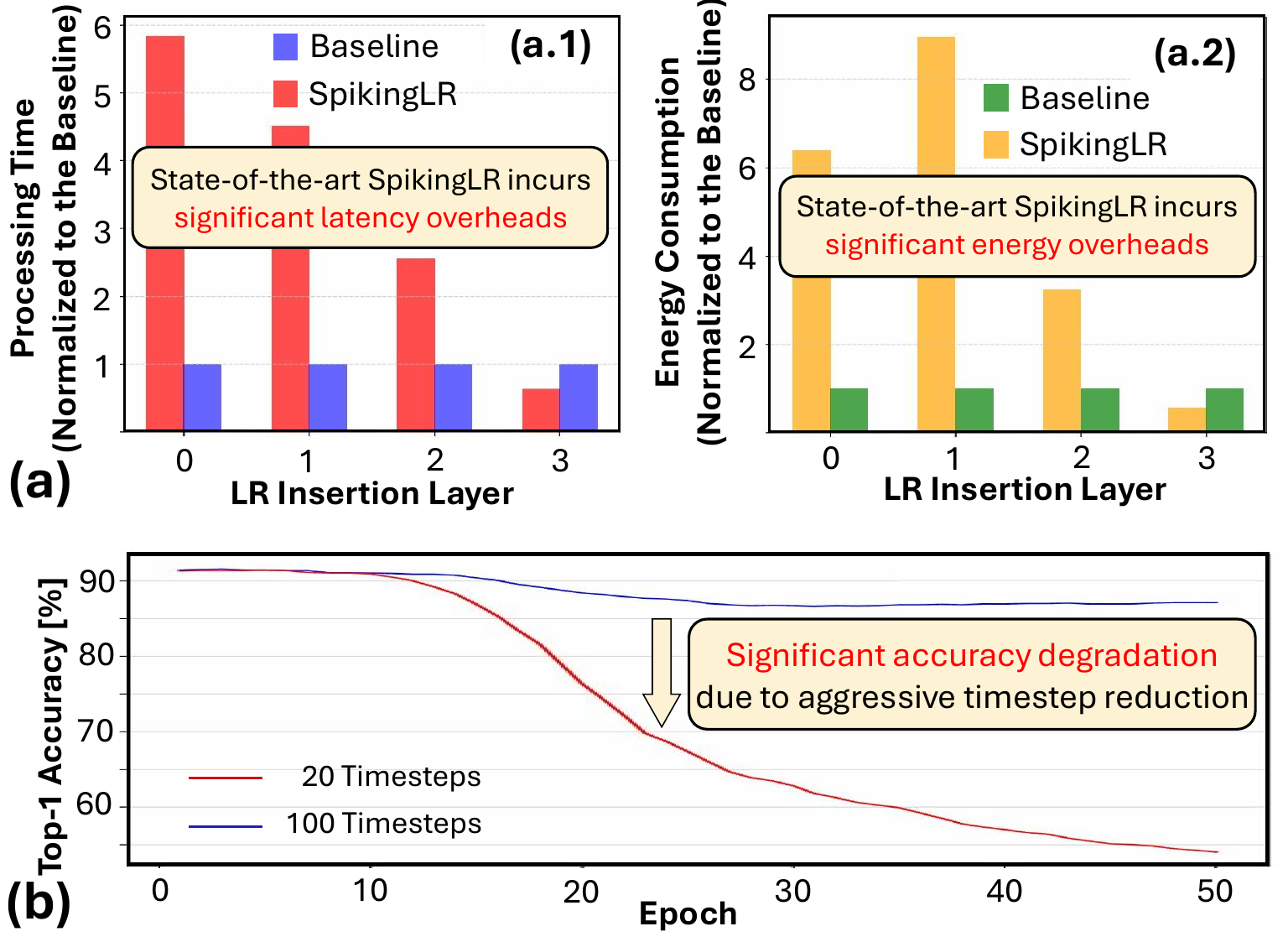}
\vspace{-0.7cm}
\caption{(a) The state-of-the-art memory replay-based work, SpikingLR~\cite{bib302}, incurs significant latency and energy overheads as compared to the baseline network without NCL techniques. (b) An aggressive timestep reduction (e.g., from 100 to 20 timesteps) may lead to significant accuracy degradation.}
\label{Fig_ObserveLimits}
\vspace{-0.5cm}
\end{figure}

In this case study, we investigate the impact of a widely-used optimization technique for SNNs (i.e., timestep reduction~\cite{Ref_Putra_TopSpark_IROS23}), while employing the memory replay-based technique in the class-incremental scenario.
To do this, we consider the implementation of SpikingLR~\cite{bib302}, and perform a timestep reduction accordingly (e.g., from 100 to 20 timesteps).
Details of the experimental setup are provided in Section~\ref{Sec_EvalMethod}.
The experimental results are presented in Fig.~\ref{Fig_ObserveLimits}, from which we identify the following key observations. 
\begin{itemize}[leftmargin=*]
    \item The state-of-the-art Spiking-LR work introduces huge latency and energy overheads due to its long timesteps.
    \item The timestep reduction may degrade the accuracy significantly, thereby hindering the implementation of optimizations on memory replay-based techniques for NCL.
\end{itemize}
Furthermore, this observation exposes several research challenges that need to be addressed, as follows.
\begin{itemize}[leftmargin=*]
    \item The processing latency should be optimized carefully so that the system can quickly respond properly to the given inputs, without a significant accuracy degradation.
    \item The enhancement techniques are required to compensate the loss of information due to the reduced timestep.  
\end{itemize}

%%%%%%%%%%%%%%%%%%%%%%%%%
\subsection{Our Novel Contributions}

\begin{figure*}[t]
    \centering
    \includegraphics[width=0.95\linewidth]{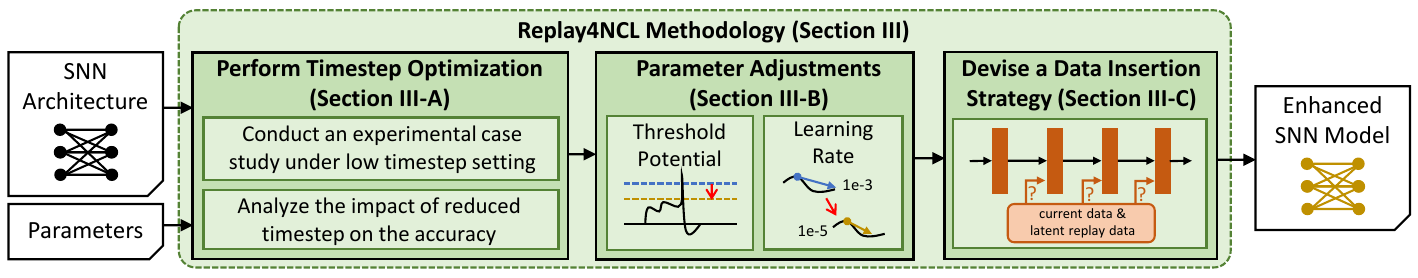}
    \vspace{-0.3cm}
    \caption{Overview of our Replay4NCL methodology, showing its key steps such as timestep optimization, parameter adjustments, and data insertion strategy; the novel contributions are highlighted in green.}
    \label{Fig_Replay4NCL}
    \vspace{-0.5cm}
\end{figure*}

To address the above challenges, we propose \textit{\textbf{Replay4NCL}, a novel methodology to enable efficient memory Replay-based NCL in resource-constrained embedded AI systems}. 
To achieve this, our Replay4NCL employs the following novel key design steps; see an overview in Fig.~\ref{Fig_Replay4NCL}.
\begin{itemize}[leftmargin=*]
    \item \textbf{Perform Timestep Optimization (Section~\ref{Sec_Replay4NCL_Timestep}):}
    It aims to introduce a low timestep setting into the SNN model to substantially reduce the processing latency and minimize the latent data memory.
    \item \textbf{Perform Parameter Adjustments (Section~\ref{Sec_Replay4NCL_Param}):}
    It aims to compensate the loss of information due to the timestep reduction, by adjusting the parameters (e.g., neuron threshold potential and learning rate) to adapt to fewer spikes in the old (latent) data. 
    \item \textbf{Devise a Data Insertion Strategy (Section~\ref{Sec_Replay4NCL_Insertion}):}
    It aims to define the layer that will receive the latent data, so that it can effectively lead to high learning quality of latent data, and hence avoiding the CF problem.
\end{itemize}

\textbf{Key Results:}
Our proposed methodology is evaluated using a Python implementation that runs on the Nvidia RTX 4090 Ti GPU machines, while considering the SHD dataset~\cite{Ref_Cramer_SHD_TNNLS20}.
The experimental results on the class-incremental scenario show that our Replay4NCL maintains old knowledge with Top-1 accuracy of 90.43\% compared to 86.22\% from the state-of-the-art SpikingLR, while effectively learning new tasks, improving the latency by 4.88x, saving latent memory by 20\%, and saving energy consumption by 36.43\%.

\begin{figure}[h]
\vspace{-0.2cm}
\centering
\includegraphics[width=\linewidth]{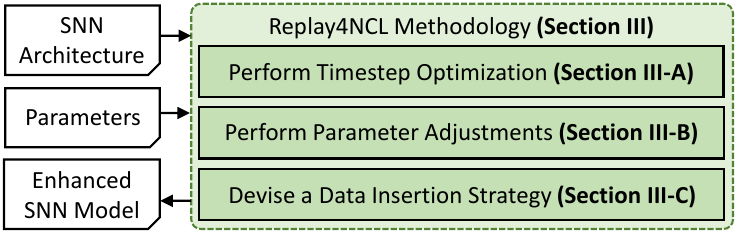}
\vspace{-0.6cm}
\caption{An overview of our novel contributions; highlighted in green.}
\label{Fig_Novelty}
\vspace{-0.2cm}
\end{figure}

%%%%%%%%%%%%%%%%%%%%%%%%%%%%%%%%%%%%%%%%%%%%%%%%%%%%%%%%%%%%%%%%%%%%%%%%
\section{Related Work}

%%%%%%%%%%%%%%%%%%%%%%%%%
\subsection{Spiking Neural Networks (SNNs)}

SNNs are the emerging neural network (NN) models, that utilize action potentials (i.e., spikes) to convey information and perform computation. 
They offer ultra-low-power/energy consumption due to their sparse spike-based operations, while achieving high performance (e.g., accuracy)~\cite{Ref_Tavanaei_DLSNN_Neunet18, Ref_Nunes_SNNsurvey_Access22, Ref_Putra_FSpiNN_TCAD20}.
SNN computations typically rely on a specific neuron model. 
The widely-used model is the Leaky Integrate-and-Fire (LIF) neuron~\cite{Ref_Izhikevich_CompareModels_TNN04}, whose internal dynamics can be expressed as follows.
\begin{equation}
    \tau \frac{dV_{mem}(t)}{dt} = - (V_{mem}(t)-V_{rst}) + Z(t) 
    \label{Eq_LIF}
\end{equation}
\begin{equation}
    \text{if} \;\;\; V_{mem} \geq V_{thr} \;\;\; \text{then} \;\;\; V_{mem} \leftarrow V_{rst}
    \label{Eq_LIF_fire}
\end{equation}
$V_{mem}$ is the neuron membrane potential at timestep-$t$, $V_{rst}$ is the neuron reset potential, and $\tau$ is to the time constant of $V_{mem}$ decay. 
Meanwhile, $Z$ is the input, and $V_{thr}$ is the neuron threshold potential.
SNNs have advantages for their ultra-low power/energy consumption and temporal information processing by leveraging event-based data, and hence the SNN deployments on tightly-constrained embedded AI systems is highly desired.

%%%%%%%%%%%%%%%%%%%%%%%%%
\subsection{Training SNNs with Surrogate Gradient (SG) Learning}

\begin{figure}[h]
    \vspace{-0.2cm}
    \centering
    \includegraphics[width=\linewidth]{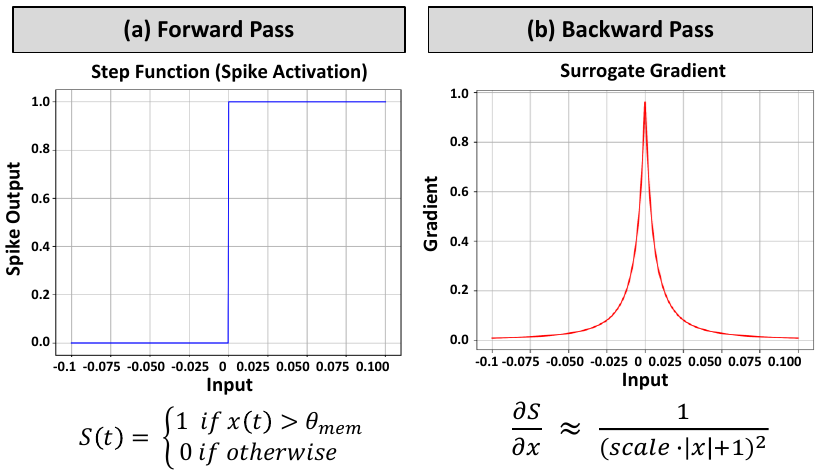}
    \vspace{-0.7cm}
    \caption{Overview of (a) the forward pass, and (b) the backward pass with SG learning technique under supervised settings.}
    \label{Fig_BPTT}
    \vspace{-0.3cm}
\end{figure}

The state-of-the-art SNN training under supervised settings typically involve the SG-based technique~\cite{Ref_Neftci_SurrogateSNNs_MSP19}\cite{Ref_Wu_STBP_FNINS18} due to its efficacy in minimizing the loss function, e.g., using the Back-Propagation Through Time (BPTT)~\cite{Ref_Werbos_BPPT_IEEE90}. 
Therefore, in this work, we also train SNNs using the SG-based BPTT by using a fast-sigmoid surrogate gradient to approximate the step function, following the studies in~\cite{bib302}; see Fig.~\ref{Fig_BPTT}. 

%%%%%%%%%%%%%%%%%%%%%%%%%
\subsection{Memory Replay-based Techniques for NCL}

\begin{figure}[t]
    \centering
    \includegraphics[width=\linewidth]{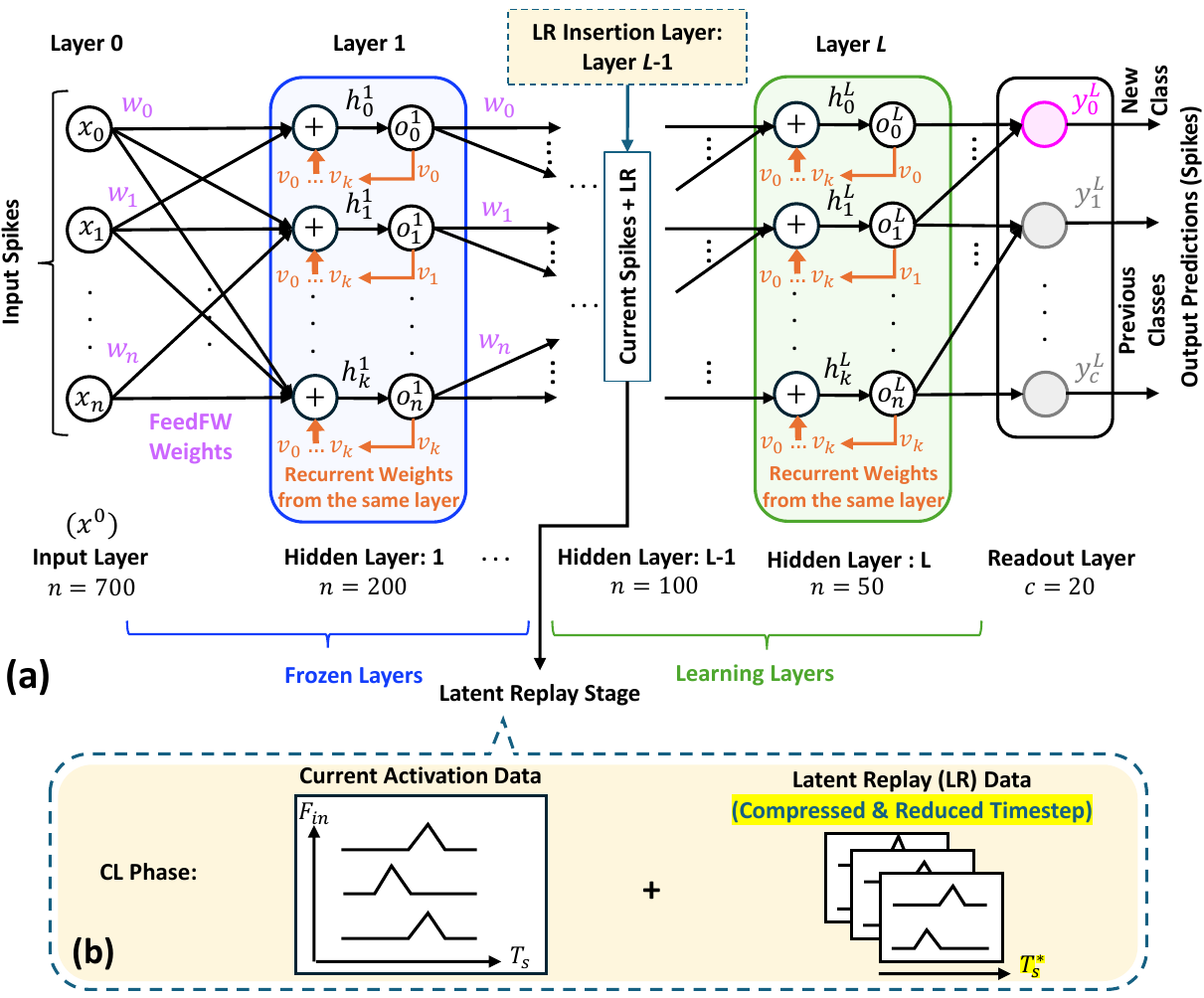}
    \vspace{-0.6cm}
    \caption{Overview of (a) SNN architecture considered in this work, and (b) configuration of current activation data and latent replay (LR) data.}
    \label{Fig_SNNarch}
    \vspace{-0.4cm}
\end{figure}

The idea of these techniques is the generation of synthetic data samples of previously learned tasks, which will be re-introduced during the NCL training phase for updating the knowledge of the system~\cite{bib19}\cite{Ref_Minhas_SurveyNCL_arXiv24}. 
This approach presents new and old tasks to the system, thereby enforcing the system to re-learn old tasks (old knowledge) through the \textit{latent replay (LR) data} to avoid CF issues, while also learning new tasks (new knowledge). 
To enable NCL with memory replay, we consider employing the SNN architecture from the SpikingLR work~\cite{bib302}, which is illustrated in Fig.~\ref{Fig_SNNarch}. 
The network comprises of \textit{frozen} and \textit{learning layers}, based on where the LR data is inserted into the network.
For instance, if we have \textit{L} layers in the network and the LR data is inserted in layer \textit{L}-1, then layer 1 until layer \textit{L}-2 are the frozen layers, while layer \textit{L}-1 and \textit{L} are the learning layers.
During the training on new data, only the learning layers are updated, while the frozen layers remain unchanged. 
These frozen layers simply forward the input spikes by processing them based on the pre-training phase.
In this work, we consider a 4-layer SNN (i.e., \textit{L} = 4).

%%%%%%%%%%%%%%%%%%%%%%%%%%%%%%%%%%%%%%%%%%%%%%%%%%%%%%%%%%%%%%%%%%%%%%%%
\section{Proposed Replay4NCL Methodology}
\label{Sec_Replay4NCL}

Our proposed methodology, illustrated in Fig.~\ref{Fig_Replay4NCL}, employs several key steps, as outlined below to enable efficient memory replay-based NCL.  

%%%%%%%%%%%%%%%%%%%%%%%%%%%%%%%%%
\subsection{Perform Timestep Optimization}
\label{Sec_Replay4NCL_Timestep}

\begin{figure}[t]
    \centering
    \includegraphics[width=0.7\linewidth]{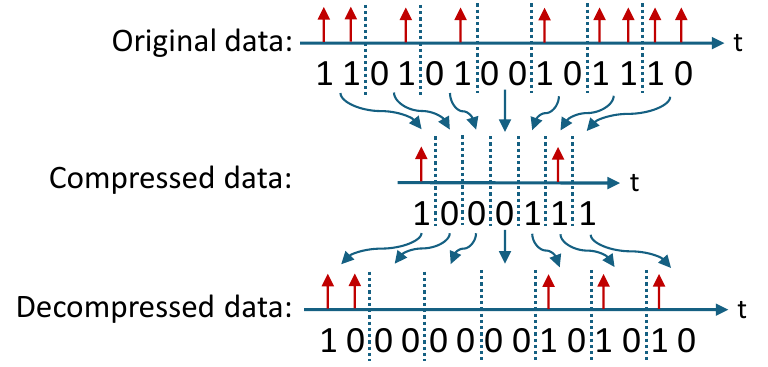}
    \vspace{-0.3cm}
    \caption{A compression and decompression mechanism considered in this work; adopted from studies in~\cite{bib302}.}
    \label{Fig_Compression}
    \vspace{-0.2cm}
\end{figure}

This step aims at optimizing the timestep setting for SNN processing, thereby substantially reducing the processing latency. 
To effectively perform such an optimization, we conduct an experimental case study that considers different timestep settings, and then analyze their impact on accuracy and latency. 
This way, we can identify the appropriate timestep settings that can provide high accuracy without any enhancements.
Furthermore, to minimize the memory size of LR data (i.e., latent memory), we also employ a data compression-decompression mechanism based on the reduced timestep (see Fig.~\ref{Fig_Compression}), thereby optimizing both the timestep and memory requirements while preserving information for NCL training phase.   

\begin{figure}[t]
    \centering
    \includegraphics[width=0.92\linewidth]{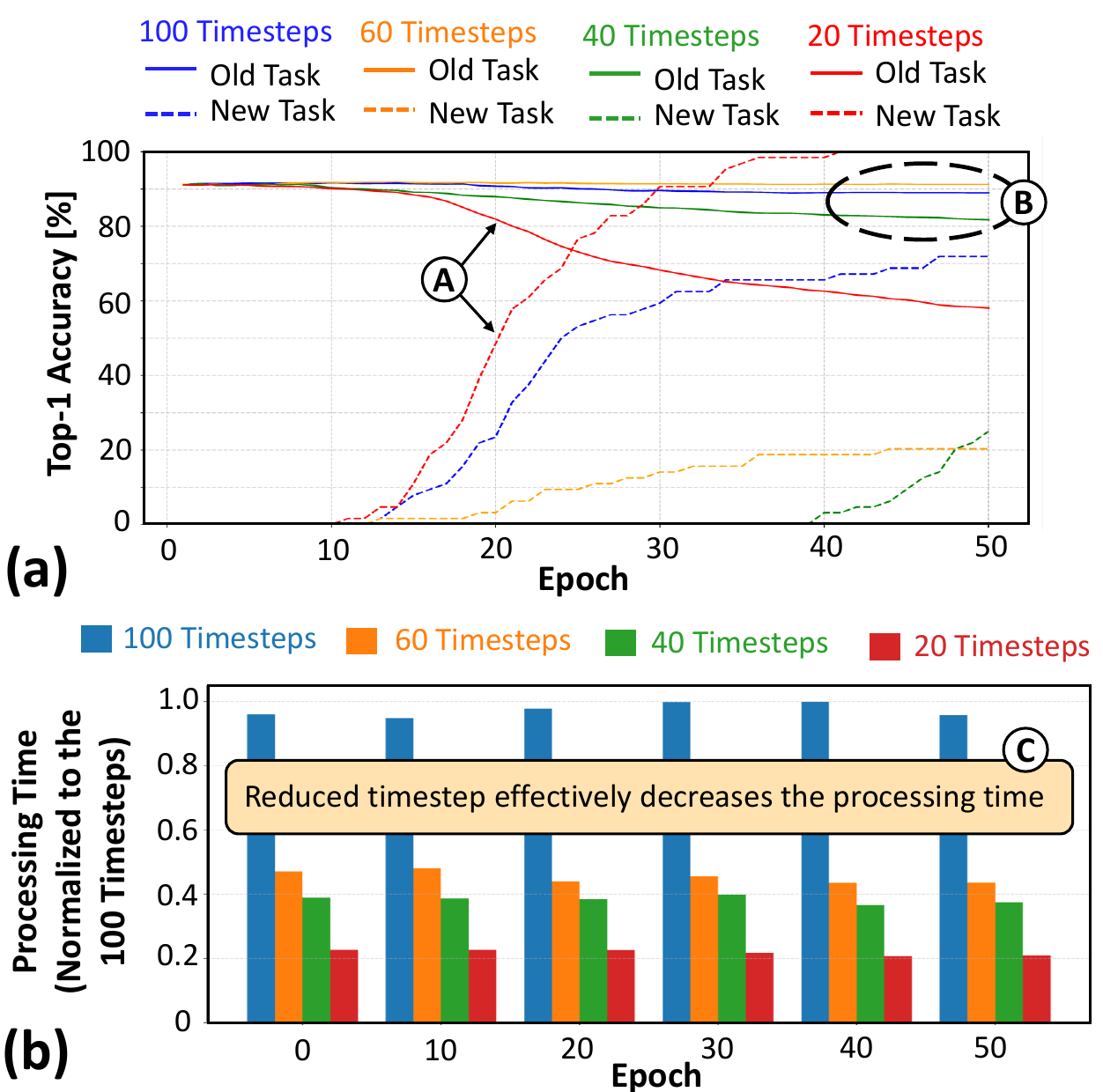}
    \vspace{-0.3cm}
    \caption{Profiles of (a) accuracy and (b) processing latency across training epochs for different timestep settings (i.e., 100, 60, 40, and 20). Here, the timestep=100 represents the setting used by the state-of-the-art~\cite{bib302}.}
    \label{Fig_ReducedTimstep}
    \vspace{-0.4cm}
\end{figure}

The experimental results are provided in Fig.~\ref{Fig_ReducedTimstep}, from which we draw several key observations below. 
\begin{itemize}[leftmargin=*]
    \item Observation \circled{A}: An aggressive timestep reduction from 100 to 20 timesteps significantly reduces the accuracy of old tasks (old knowledge). 
    \item Observation \circled{B}: To obtain acceptable accuracy for the old tasks (old knowledge) comparable to the state-of-the-art, we should employ at least 40 timesteps.
    \item Observation \circled{C}: Reduced timestep effectively decreases the processing time (latency), but they should be leveraged carefully to avoid significant accuracy degradation.
\end{itemize}
These observation points are leveraged for devising effective parameter adjustments in Section~\ref{Sec_Replay4NCL_Param}.

%%%%%%%%%%%%%%%%%%%%%%%%%%%%%%%%%
\subsection{Parameter Adjustments}
\label{Sec_Replay4NCL_Param}

This step aims at compensating the information loss due to the reduced timestep. 
Here, we observe that the generated spikes are less as compared to the original pre-trained model.
The reason is that, neurons still employ the same threshold potential $V_{thr}$ like the pre-trained model, but they encounter fewer number of spikes, which may not be enough for increasing the membrane potential $V_{mem}$ to reach the threshold $V_{thr}$.
Therefore, to make the network model adapt better to fewer spikes, we adjust the values of threshold potential $V_{thr}$ and learning rate $\eta$ with the following policy.
\begin{itemize}[leftmargin=*]
    \item \textbf{Threshold Potential ($V_{thr}$):}
    We adjust $V_{thr}$ by reducing its value, since it will decrease the number of incoming spikes required to make $V_{mem}$ reach $V_{thr}$, and hence preserving the spiking activity closer to the original pre-trained model.
    \item \textbf{Learning Rate ($\eta$):}
    We adjust $\eta$ also by reducing its value, as it will decrease the learning speed of the network. 
    This will ensure more careful weight updates during the training phase, considering that the network needs to leverage fewer spikes compared to the original pre-trained model. 
\end{itemize}

%%%%%%%%%%%%%%%%%%%%%%%%%%%%%%%%%
\subsection{Devise a Latent Replay Data Insertion Strategy}
\label{Sec_Replay4NCL_Insertion}

This step aims at devising a strategy to effectively benefit from the LR data insertion into the network.
To achieve this, we perform a design exploration to see the impact of LR data insertion into different layers, and select the one that offers the highest accuracy. 
Additionally, this exploration also includes timestep optimization and parameter adjustments from Sections~\ref{Sec_Replay4NCL_Timestep} and~\ref{Sec_Replay4NCL_Param}. 
To incorporate all these techniques into an integrated mechanism, we propose a training strategy, whose key steps are presented in Alg.~\ref{Alg_Strategy} and discussed below.
\begin{itemize}[leftmargin=*]
    \item First, we perform the pre-training phase, where the SNN model is trained and developed to learn all tasks from pre-training dataset; see Alg.~\ref{Alg_Strategy}: line 1-5.
    \item Second, we prepare the network for NCL training phase. 
    To do this, we perform the following. 
    \begin{itemize}
        \item We generate the LR data activations ($A_{LR}$), and then split the SNN model based on the selected layer for LR data insertion (i.e., so-called \textit{LR layer}), following the approach from~\cite{bib302}; see Alg.~\ref{Alg_Strategy}: line 6-20.
        \item We propose to dynamically change the threshold potential $V_{thr}$ based on the timestep, adjustment interval, and spike timing information.
        If the spikes occur during the defined interval, $V_{thr}$ is increased; otherwise $V_{thr}$ is decreased using a sigmoidal decay; see Alg.~\ref{Alg_Strategy}: line 10-17.
        This way, the generated $A_{LR}$ adapt better to the reduced timestep.
    \end{itemize}
    \item Third, we perform the NCL training phase by learning new tasks, while considering the following actions.
    \begin{itemize}
        \item We dynamically adjust $V_{thr}$ based on the timestep, adjustment interval, and spike timing information. 
        This adaptive $V_{thr}$ is updated based on the spike timing. 
        If the spikes occur during the defined interval, $V_{thr}$ is increased; otherwise $V_{thr}$ is decreased; see Alg.~\ref{Alg_Strategy}: line 25-30.
        \item We also reduce the learning rate $\eta$ to make the network learn from fewer spikes more carefully; see Alg.~\ref{Alg_Strategy}: 21-33. 
    \end{itemize}
    As a result, the trained network is considered as the final output; see Alg.~\ref{Alg_Strategy}: 34-35.
\end{itemize}

%%%-----
\begin{algorithm}[t]
  \footnotesize
  \caption{LR Data Training Strategy}
  \label{Alg_Strategy}
  \begin{algorithmic}[1]
	\REQUIRE 
        \textbf{[1]} Training set: pre-training ($TS_{pre}$), continual learning ($TS_{cl}$), subset of the pre-training ($TS_{replay}$); \\
        \textbf{[2]} Network parameters: network ($net$), number of layers ($L$), threshold potential ($V_{thr}$), learning rate ($\eta$), insertion layer for LR data ($L_{ins}$), reduced timesteps ($T_{step}$), adjust\_interval=5;\\
        \textbf{[3]} Epoch: pre-training ($E_{pre}$), continual learning ($E_{cl}$); \\
        \textbf{[4]} Activations: current ($A_{new}$), LR ($A_{LR}$); \\
	\ENSURE 
        Enhanced SNN model ($SNN$); \\
        \smallskip
	\textbf{BEGIN} \\
	  \textbf{Pre-training phase}: \\
	  \STATE Initialize SNN weights \textit{w} and \textit{v} randomly; \\ 
        \STATE $\eta_{pre} =$ 1e-3;
        \FOR{($ep$=0; $ep<E_{pre}$; $ep$++)}
          \STATE $w,v \leftarrow$ train$_{pre}$($net$, ($w, v$), $TS_{pre}$, $\eta_{pre}$, $V_{th}$);
        \ENDFOR\\
        \smallskip
	  \textbf{Network preparation for NCL training phase}: \\
        \STATE $\eta_{cl} = \eta_{pre}/100$; \\
	  \STATE $net_{f}, net_{l} =$ split($net, L_{ins}$); \\ 
        \FOR{($ep$=0; $ep<E_{pre}$; $ep$++)}
          \FOR{($t$=0; $t<T_{step}$; $t$++)}
            \IF{($t$ \% adjust\_interval == 0)}
              \IF{any spikes occur}
                \STATE $avg\_spike\_time =$ mean($spike\_timing$);
                \STATE $V_{thr}=$ 1 + 0.01($T_{step}-avg\_spike\_time$); \\
              \ENDIF \\
            \ELSE
              \STATE $V_{thr}=$ 1 / (1 + exp(-0.001 $\cdot$ $t$)); \\
            \ENDIF\\
             \STATE $A_{LR} =$ infer($net_{f}$, $TS_{replay} \subseteq TS_{pre}$); \\
          \ENDFOR\\
        \ENDFOR\\
        \smallskip
	  \textbf{NCL training phase}: \\
        \STATE $\eta_{cl} = \eta_{pre}/100$; \\
        \FOR{($ep$=0; $ep<E_{cl}$; $ep$++)}
          \STATE $A_{new} =$ inference ($net_{f}$, $TS_{cl}$); \\
          \FOR{($t$=0; $t<T_{step}$; $t$++)}
            \IF{any spikes occur}
                \STATE $avg\_spike\_time =$ mean($spike\_timing$);
                \STATE $V_{thr}=$ 1 + 0.01($T_{step}-avg\_spike\_time$); \\
            \ELSE
              \STATE $V_{thr}=$ 1 / (1 + exp(-0.001 $\cdot$ $t$)); \\
            \ENDIF\\
            \STATE $w_l, v_l \leftarrow$ train($net_l$, ($w_l, v_l$), $A_{new} \cup A_{LR}$, $\eta_{cl}$, $V_{thr}$); \\
          \ENDFOR\\
        \ENDFOR\\
        \STATE $SNN \leftarrow net$; \\
        \smallskip
	\RETURN $SNN$; \\
	\textbf{END}
	\end{algorithmic}
\end{algorithm}
\setlength{\textfloatsep}{8pt}

%%%%%%%%%%%%%%%%%%%%%%%%%%%%%%%%%%%%%%%%%%%%%%%%%%%%%%%%%%%%%%%%%%%%%%%%
\section{Evaluation Methodology}
\label{Sec_EvalMethod}

\begin{figure}[h]
    \vspace{-0.2cm}
    \centering
    \includegraphics[width=0.8\linewidth]{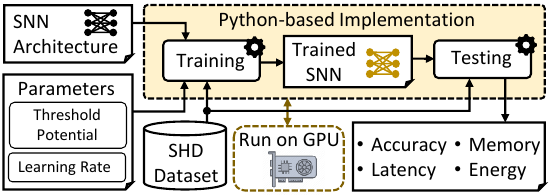}
    \vspace{-0.3cm}
    \caption{Experimental setup considered in this work.}
    \label{Fig_EvalMethod}
    \vspace{-0.2cm}
\end{figure}

Fig.~\ref{Fig_EvalMethod} illustrates the experimental setup for evaluating our
Replay4NCL methodology.
Here, we employ a Python-based implementation that runs on the Nvidia RTX 4090 Ti GPU machines with the SHD dataset~\cite{Ref_Cramer_SHD_TNNLS20} as the workload. 
For the network architecture, we use the SNN architecture shown in Fig.~\ref{Fig_SNNarch} with 4 layers.
We also consider the class-incremental scenario, i.e., the SNN is trained with 19 tasks (classes) during the pre-training phase, and then it is trained with 20$^{th}$ task (class) during the CL training phase. 
For the comparison partner, we consider the SpikingLR with its compressed LR data technique~\cite{bib302} as the state-of-the-art.
The evaluation generates several results, including the accuracy for new and old tasks, latency, latent memory size, and energy consumption. 

%%%%%%%%%%%%%%%%%%%%%%%%%%%%%%%%%%%%%%%%%%%%%%%%%%%%%%%%%%%%%%%%%%%%%%%%
\vspace{-0.1cm}
\section{Experimental Results and Discussion}
\label{Sec_Results}

\begin{figure*}[t]
    \centering
    \includegraphics[width=0.86\linewidth]{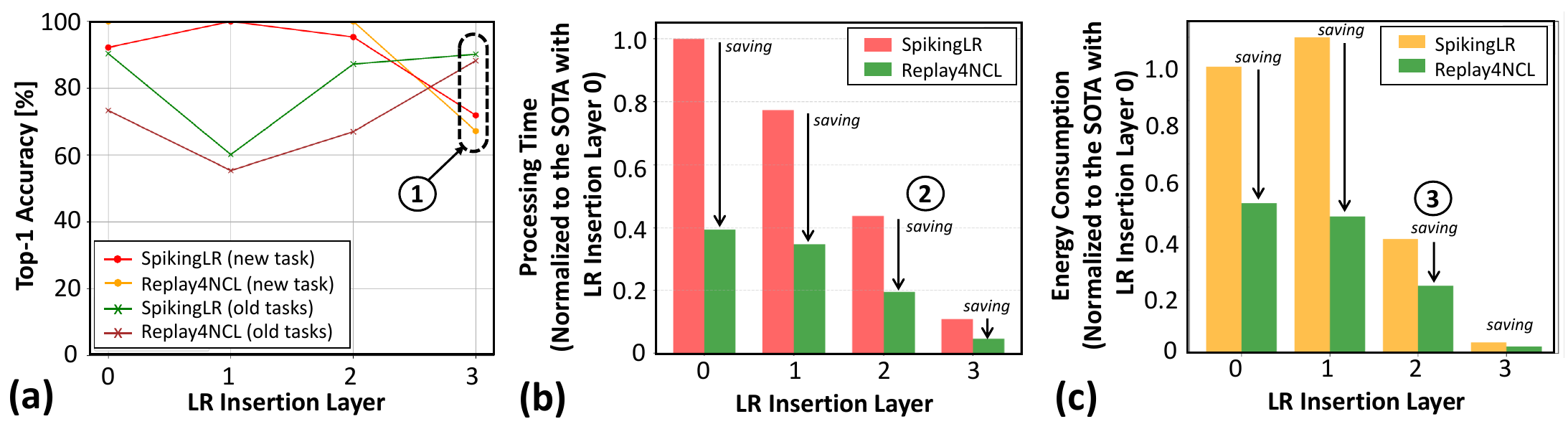}
    \vspace{-0.3cm}
    \caption{Results for (a) Top-1 accuracy considering both old and new tasks, (b) processing time, and (c) energy consumption of the SpikingLR and our Replay4NCL across different LR insertion layers.}
    \label{Fig_Results_AcrossLayers}
    \vspace{-0.3cm}
\end{figure*}

\begin{figure*}[t]
    \centering
    \includegraphics[width=0.75\linewidth]{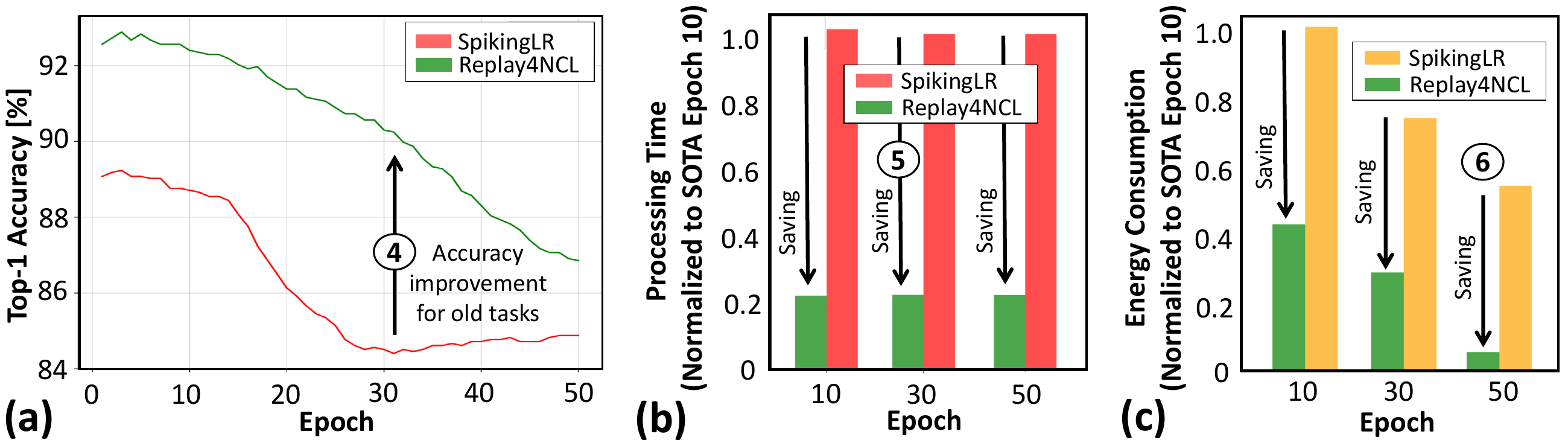}
    \vspace{-0.3cm}
    \caption{Results for (a) Top-1 accuracy of old tasks, (b) processing time, and (c) energy consumption of the SpikingLR and our Replay4NCL across different training epochs considering the LR insertion layer 3.}
    \label{Fig_Results_Layer3}
    \vspace{-0.5cm}
\end{figure*}

\begin{figure}[t]
    \centering
    \includegraphics[width=\linewidth]{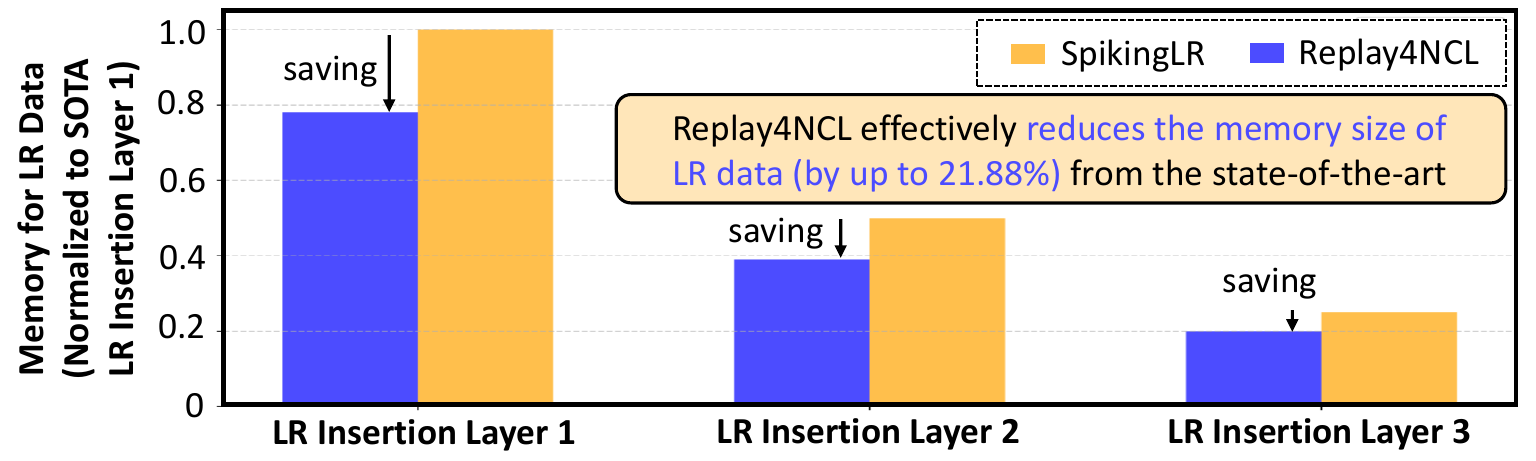}
    \vspace{-0.6cm}
    \caption{Comparison of the latent memory sizes between the SpikingLR and our Replay4NCL across different LR insertion layers.}
    \label{Fig_Results_Memory}
    \vspace{-0.1cm}
\end{figure}

%%%%%%%%%%%%%%%%%%%%%%%%%%%%%%%%%
\subsection{Maintaining High Accuracy for Old and New Tasks}
\label{Sec_Results_Accuracy}

Experimental results for accuracy across different LR insertion layers are shown in Fig.~\ref{Fig_Results_AcrossLayers}(a). 
In general, we observe that our Replay4NCL achieves comparable accuracy to the SpikingLR across different LR insertion layers for learning a new task. 
Moreover, for the LR insertion layers 0-2, our Replay4NCL achieves 100\% accuracy when learning a new task within 50 epochs, which is better than the SpikingLR.
For the LR insertion layer 3, the Replay4NCL achieves comparable accuracy to the SpikingLR when learning a new task (i.e., 67.19\% for Replay4NCL and 71.88\% for the SpikingLR); see \circled{1}.
Meanwhile, when learning old tasks, our Replay4NCL achieves comparable high accuracy to the SpikingLR in the LR insertion layer 3 (i.e., 88.24\% for Replay4NCL and 90.19\% for the state-of-the-art); see \circled{1}.
These results indicate that the Replay4NCL maintains high accuracy comparable to the SpikingLR on both learning the old and new tasks, despite employing low timestep settings.
The reason is the Replay4NCL employs parameter adjustments to compensate the information loss due to fewer spikes, and leverages a training strategy that determines the effective threshold potential $V_{thr}$ and learning rate $\eta$ values.   

Meanwhile, experimental results for accuracy across different training epochs considering the LR insertion layer 3, are provided in Fig.~\ref{Fig_Results_Layer3}(a). 
We observe that Replay4NCL achieves higher Top-1 accuracy than the SpikingLR (see \circled{4}), since its parameter adjustments and training strategy effectively improve the learning process on the later layers.

%%%%%%%%%%%%%%%%%%%%%%%%%%%%%%%%%
\subsection{Improvements on Processing Latency}
\label{Sec_Results_Latency}

Experimental results for processing time (latency) across different LR insertion layers are provided in Fig.~\ref{Fig_Results_AcrossLayers}(b).
In general, our Replay4NCL significantly reduces the processing time as compared to the SpikingLR, i.e., by up to 2.34x speed-up; see \circled{2}.
Similar trends are also observed for processing latency in the LR insertion layer 3 case, as shown by \circled{5} in Fig.~\ref{Fig_Results_Layer3}(b).
These improvements are obtained mainly due to the reduction of timestep settings.

%%%%%%%%%%%%%%%%%%%%%%%%%%%%%%%%%
\subsection{Latent Memory Savings}
\label{Sec_Results_Memory}

Experimental results for the latent memory size considering different LR insertion layers are provided in Fig.~\ref{Fig_Results_Memory}. 
In general, we observe that later layers tend to require smaller latent memory sizes due to smaller layer dimension (i.e., fewer number of neurons). 
Furthermore, these results also show that our Replay4NCL significantly saves the latent memory size as compared to the SpikingLR by 20\%-21.88\%, due to the timestep reduction which decreases the size of spike-based LR data activations to be stored in the memory.
Such memory savings are beneficial for ensuring the applicability of our Replay4NCL for tightly-constrained embedded AI systems.

%%%%%%%%%%%%%%%%%%%%%%%%%%%%%%%%%
\subsection{Reduction on Energy Consumption}
\label{Sec_Results_Energy}

Experimental results for energy consumption across different LR insertion layers are provided in Fig.~\ref{Fig_Results_AcrossLayers}(c). 
In general, we observe that our Replay4NCL significantly decreases the energy consumption from the SpikingLR across different LR insertion layers. 
Specifically, the Replay4NCL offers up to 56.7\% energy savings; see \circled{3}.
Similar trends are also observed for energy consumption in the LR insertion layer 3, as the Replay4NCL offers 36.4\% energy saving; see \circled{6} in Fig.~\ref{Fig_Results_Layer3}(c).
The reason is that, our Replay4NCL employs short timestep settings which significantly reduce the processing latency, thereby leading to low energy consumption.

%%%%%%%%%%%%%%%%%%%%%%%%%%%%%%%%%
\subsection{Further Discussion}
\label{Sec_Results_Discuss}

\begin{figure}[t]
    \centering
    \includegraphics[width=0.98\linewidth]{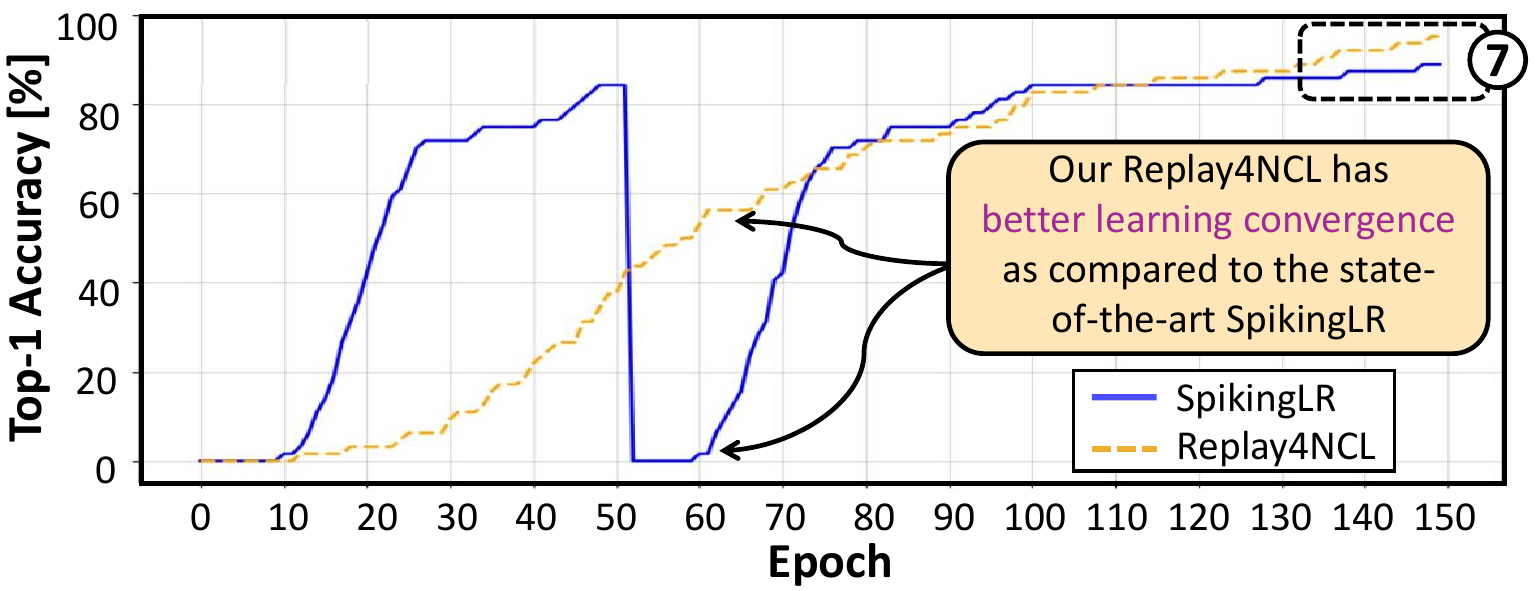}
    \vspace{-0.3cm}
    \caption{Comparison of accuracy profiles between the SpikingLR and our Replay4NCL when learning a new task, while considering a long training period, i.e., 150 training epochs.}
    \label{Fig_Results_LongEpoch}
\end{figure}

We further investigate the accuracy of the SpikingLR and our Replay4NCL considering a long training period (i.e., 150 epochs), and the experimental results are shown in Fig.~\ref{Fig_Results_LongEpoch}.
These results show that our Replay4NCL can achieve higher accuracy than the SpikingLR, as shown by \circled{7}. 
This indicates that the proposed LR training strategy in Replay4NCL provides an effective integrated mechanism to maximize the benefits from the proposed optimization through timestep reduction, and the NCL training enhancements through parameter adjustments. 
Moreover, with the employment of a lower learning rate $\eta$ than the SpikingLR, our Replay4NCL offers more careful weight updates, which in turns leading to better learning convergence when processing fewer spikes as indicated by a smoother learning curve. 

%%%%%%%%%%%%%%%%%%%%%%%%%%%%%%%%%%%%%%%%%%%%%%%%%%%%%%%%%%%%%%%%%%%%%%%%
\section{Conclusion}
\label{Sec_Conclusion}

In this paper, we propose the Replay4NCL methodology to enable efficient memory replay for NCL in embedded AI systems. 
Our Replay4NCL compresses LR data (old knowledge) and replays them during the CL training phase using small timestep settings, hence reducing latency and energy consumption. 
To compensate the information loss due to fewer spikes, we adapt the neuron threshold potential and learning rate settings. 
Experimental results on the class-incremental scenario with the SHD dataset demonstrate that Replay4NCL effectively supports the new task learning, while avoiding CF problem.
Specifically, Replay4NCL maintains old knowledge with Top-1 accuracy of 90.43\% compared to the state-of-the-art with 86.22\%, while improving the latency by 4.88x, saving the latent memory by 20\%, and saving the energy consumption by 36.43\%.
These results demonstrate the potential of our Replay4NCL methodology for enabling NCL for tightly-constrained embedded AI systems.

\vspace{-0.1cm}
%%%%%%%%%%%%%%%%%%%%%%%%%%%%%%%%%%%%%%%%%%%%%%%%%%%%%%%%%%%%%%%%%%%%%%%%
\section*{Acknowledgment}
This work was partially supported by United Arab Emirates University (UPAR) with Fund code 12N169. This work was also partially supported by the NYUAD Center for Artificial Intelligence and Robotics (CAIR), funded by Tamkeen under the NYUAD Research Institute Award CG010.

%%%%%%%%%%%%%%%%%%%%%%%%%%%%%%%%%%%%%%%%%%%%%%%%%%%%%%%%%%%%%%%%%%%%%%%%
\bibliographystyle{IEEEtran}
\bibliography{bibliography}

\end{document}